\documentclass{article}

\usepackage{microtype}
\usepackage{graphicx}
\usepackage{subcaption}
\usepackage{booktabs} 

\usepackage{hyperref}

\let\origaddcontentsline\addcontentsline %

\usepackage[preprint]{icml2026}

\usepackage{amsmath}
\usepackage{amssymb}
\usepackage{mathtools}
\usepackage{amsthm}

\usepackage[capitalize,noabbrev]{cleveref}

\theoremstyle{plain}

\theoremstyle{definition}

\theoremstyle{remark}

\usepackage[textsize=tiny]{todonotes}

\icmltitlerunning{Laplacian In-context Spectral Analysis}

\usepackage{rotating}
\usepackage{mathrsfs}

\newcommand{\Softmax}{\mathrm{Softmax}}
\newcommand{\softmax}{\mathrm{softmax}}

\let\addcontentsline\origaddcontentsline
\usepackage{amsmath, amsthm}        
\usepackage{extarrows}      
\usepackage{mathrsfs}       
\usepackage{tikz}           
\usetikzlibrary{shapes.geometric, arrows, calc, fadings, decorations.pathreplacing} 

\tikzset{%
  >=latex, 
  inner sep=0pt,%
  outer sep=2pt,%
  mark coordinate/.style={inner sep=0pt,outer sep=0pt,minimum size=3pt,
    fill=black,circle}%
}
\usetikzlibrary{arrows.meta,positioning,shapes.geometric}

\begin{document}

\twocolumn[
  \icmltitle{LISA: Laplacian In-context Spectral Analysis}



  \icmlsetsymbol{equal}{*}

  \begin{icmlauthorlist}
    \icmlauthor{Julio Candañedo}{yyy}
  \end{icmlauthorlist}

  \icmlaffiliation{yyy}{SparseTrace.ai, Appleton, Wisconsin, USA}
  \icmlcorrespondingauthor{Julio Candanedo}{julio@sparsetrace.ai}

  \icmlkeywords{Machine Learning, ICML, Time-Series}

  \vskip 0.3in

]



\printAffiliationsAndNotice{}  

\begin{abstract}
We propose Laplacian In-context Spectral Analysis (LISA), a method for inference-time adaptation of Laplacian-based time-series models using only an observed prefix. LISA combines delay-coordinate embeddings and Laplacian spectral learning to produce diffusion-coordinate state representations, together with a frozen nonlinear decoder for one-step prediction. We introduce lightweight latent-space residual adapters based on either Gaussian-process regression or an attention-like Markov operator over context windows. Across forecasting and autoregressive rollout experiments, LISA improves over the frozen baseline and is often most beneficial under changing dynamics. This work links in-context adaptation to nonparametric spectral methods for dynamical systems.
\end{abstract}


\section{Introduction}

Many scientific datasets are \emph{continuous} sequences---from dynamical trajectories to climate and electricity demand---whose structure is only apparent when interpreted through context. Yet long-context forecasting for real-valued time series remains difficult: dynamics can be nonlinear, partially observed, noisy, and often nonstationary. In contrast, discrete sequence modeling has seen major breakthroughs, most prominently through the transformer architecture~\cite{vaswani2017attention}, which enables scalable attention-based modeling over long contexts. Subsequent work identified an inference-time phenomenon known as \emph{in-context learning} (ICL)~\cite{brown2020_fewShotLearners, dong2024_inContextLearningSurvey}, where models adapt their predictions to information in the prompt without parameter updates. This motivates us to study \emph{in-context mechanisms} (ICMs) for continuous-time-series prediction: lightweight, prompt-specific updates that exploit additional history at test time.

In this work, we propose \textit{Laplacian In-context Spectral Analysis} (LISA), a framework for endowing nonparametric geometric time-series models with an ICL-like adaptation mechanism. Moreover, temporal sequence models often exhibit a stronger \emph{continuity bias} than generic manifold learning---nearby states in time are typically constrained to evolve smoothly---making them particularly well-suited for constructing stable representations from context. Our perspective is that ICL can be understood as the interaction of an \emph{underlying parametric model} with a lightweight \emph{nonparametric adapter} that updates behavior using only context. Within LISA, we develop two such nonparametric adapters: one based on \emph{Markov operators} and one based on \emph{Gaussian processes} (GPs). Our core model is built from Laplacian-based manifold learning---Laplacian eigenmaps / diffusion maps~\cite{Belkin2003, coifman2006_diffusionMaps}---combined with delay-coordinate embeddings motivated by Takens' theorem~\cite{Takens1981}. The resulting theory falls under \emph{Nonlinear Laplacian Spectral Analysis} (NLSA)~\cite{Giannakis2012_NLSA}, providing a principled nonparametric representation of dynamical structure in continuous sequences. Building on the ideas of~\cite{candanedo2024_DMAE}, we show how this nonparametric construction can be converted into an explicit parametric model in autoencoder form, with weights that can be solved for exactly. This yields an \emph{encoder} given by NLSA diffusion coordinates and a \emph{decoder} that takes the form of a Gaussian-Process Latent Model (GPLM), closely related to the GP-LVM framework~\cite{lawrence2004_gplvm}.

\begin{figure}[!hbt]
\centering
\resizebox{\columnwidth}{!}{%
\begin{tikzpicture}[
  >=Stealth,
  node distance=11mm,
  every node/.style={font=\small},
  io/.style={inner sep=0pt},
  block/.style={draw, thick, align=center, minimum height=7mm, minimum width=23mm},
  enc/.style={block, trapezium, trapezium angle=110, shape border rotate=90},
  dec/.style={block, trapezium, trapezium angle=110, shape border rotate=270},
  arr/.style={->, thick},
  darr/.style={->, thick, dashed}
]
  \node (ctx) [io] {$R_{aX}$};
  \node (enc) [enc, right=18mm of ctx  ] {\textbf{NLSA}\\Encoder};
  \node (dec) [dec, right=16mm of enc  ] {\textbf{GPLM}\\Decoder};
  \node (icm) [block, right=16mm of dec] {\textbf{ICM}};
  \node (out) [io, right=18mm of icm] {${R}_{X}$};

  \draw[arr]  (ctx) -- (enc);
  \draw[arr]  (enc) -- node[above] {$R_{Ax}$} (dec);
  \draw[arr]  (dec) -- node[above] {$R_{AX}$} (icm);
  \draw[arr]  (icm) -- (out);
\end{tikzpicture}%
}
\caption{LISA schematic. The NLSA encoder constructs diffusion coordinates from the \emph{context} sequence via delay embedding and Laplacian spectral analysis. The in-context mechanism (ICM) performs a lightweight, prompt-specific nonparametric update (Markov-operator or GP-based), producing an adapted model instance used by the GPLM decoder to reconstruct or predict outputs for query samples.}
\label{fig:LISA_schematic}
\end{figure}
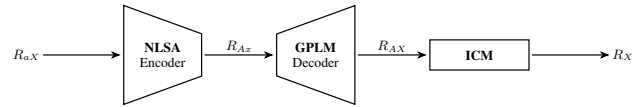

Figure~\ref{fig:LISA_schematic} summarizes the LISA architecture. Given a context sequence $R_{aX}$, the NLSA encoder constructs a nonparametric state representation $R_{ax}$ by applying delay-coordinate embeddings and Laplacian spectral analysis to the \emph{entire} context. Importantly, this encoder step is context-driven and does not require gradient-based training or online optimization. The in-context mechanism (ICM) then performs a lightweight prompt-time update using only context information, producing an adapted model instance (e.g., a Markov operator or GP posterior) that modifies downstream behavior without changing global parameters. Finally, the GPLM decoder maps the latent representation back to the ambient space to produce reconstructions or forecasts for novel query samples $R_{X}$.

\section{In-Context Mechanism \label{ICM}}
\subsection{Motivation \label{subsec:motivation}}

Suppose our training data are drawn from a Gaussian distribution $\mathcal{P} \equiv \mathcal{N}\!\left(0_X,\Sigma_{XX'}\right)$. Where $X\in\mathbb{Z}_D$ indexes ambient coordinates and $\Sigma_{XX'}$ is the population covariance. An empirical dataset is obtained by sampling $P_{iX} \sim \mathcal{P},\,\, i\in\mathbb{Z}_N.$ Let $P_X := \frac{1}{N}\sum_{i=1}^N P_{iX}$ denote the empirical mean. The empirical covariance is
\begin{align}
\Sigma^{(N)}_{XX'} := \frac{1}{N-1}\sum_{i=1}^N (P_{iX}-P_X)(P_{iX'}-P_{X'}).
\end{align}
In the large-$N$ limit, $\Sigma^{(N)} \to \Sigma$, with typical estimation error scaling as $\mathcal{O}(N^{-1/2})$. PCA on the $\mathcal{P}$-data yields a loading matrix $V^{(\mathcal{P})}_{Xx}\in\mathbb{R}^{D\times r}$ (columns are principal directions). We will use the equivalent transpose convention $V^{(\mathcal{P})}_{xX} := \left(V^{(\mathcal{P})}_{Xx}\right)^\top \in\mathbb{R}^{r\times D}$, so that a novel sample $p_X\sim\mathcal{P}$ is embedded as:
\begin{align}
p_x = V^{(\mathcal{P})}_{xX}\,p_X.
\end{align}
However, consider a distribution shift to $\mathcal{Q} \equiv \mathcal{N}\!\left(0_X,\Sigma'_{XX'}\right)$, and a sample $q_X\sim\mathcal{Q}$. Embedding $q_X$ using the \emph{training} loading matrix produces the \emph{$\mathcal{P}$-coordinates of a $\mathcal{Q}$-sample}
\begin{align}
q'_x := V^{(\mathcal{P})}_{xX}\,q_X,
\end{align}
which need not coincide with the latent coordinates $q_x$ that would be obtained by performing PCA under $\mathcal{Q}$. We therefore posit that, locally, the discrepancy between these coordinate systems is approximated by an (unknown) linear map
\begin{align}
q_x \approx M_{xx'}\,q'_{x'}.
\end{align}
Equivalently, composing $M$ with the training PCA basis yields an adapted loading matrix
\begin{align}
W_{xX} := M_{xx'}\,V^{(\mathcal{P})}_{x'X}, \qquad\text{so that}\qquad q_x \approx W_{xX}\,q_X.
\end{align}
Thus, the core in-context problem is to infer $M_{xx'}$ (or $W_{xX}$) from a short context drawn from $\mathcal{Q}$.

\paragraph{Oracle solution (known covariances).}
If $\Sigma_{XX'}$ and $\Sigma'_{XX'}$ were known, we could compute the best linear map from ambient coordinates to $\mathcal{P}$-latents (and then to $\mathcal{Q}$-latents) in closed form. Define the covariance of $\mathcal{Q}$ expressed in $\mathcal{P}$-latent coordinates,
\begin{align}
\Sigma'^{(\mathcal{P})}_{xx'} &:= V^{(\mathcal{P})}_{xX}\,\Sigma'_{XX'}\,V^{(\mathcal{P})}_{x'X'},
\end{align}
and the cross-covariance between ambient coordinates and these $\mathcal{P}$-latents,
\begin{align}
\Sigma'^{(\mathcal{P})}_{Xx} &:= \Sigma'_{XX'}\,V^{(\mathcal{P})}_{xX'}.
\end{align}
Then the optimal linear regression (in the mean-square sense) from $q'_x$ to $q_X$ is
\begin{align}
W_{Xx} = \Sigma'^{(\mathcal{P})}_{Xx} \left(\Sigma'^{(\mathcal{P})}_{xx'}\right)^{-1}_{x'x}.
\end{align}
In practice, however, $\Sigma'_{XX'}$ is unknown and must be estimated from a short context of length $\ell$, while $\Sigma_{XX'}$ and $V^{(\mathcal{P})}_{xX}$ are learned from the large training set of size $N\gg \ell$.

The IC-PCA perspective suggests that distribution shift induces a \emph{context-dependent reparameterization} of coordinates, which can be estimated from a short prefix. In our time-series setting, rather than adapting the PCA basis directly, we use the prefix to build a table of context examples and learn an \emph{in-context correction} using kernel methods (Gaussian-process / kernel ridge or Nadaraya--Watson). These mechanisms can be viewed as nonlinear generalizations of the above linear adaptation, operating on latent coordinates obtained from Hankelized windows. The picture of forecasting time-series changes the analysis because our true future latents go through the entire forecasting auto-encoder, of fig. \ref{ICM}.

\subsection{Hankelization \label{Hankelization}}

Given a prefix $R_{aX}\in\mathbb{R}^{\ell\times D}$ with $\ell\ge L$, we form $L$-delay windows by Hankelization:
\begin{align}
R_{A c X} \;:=\; \mathscr{H}(R_{aX})
\in \mathbb{R}^{(C+1)\times L \times D},
\qquad
C := \ell - L,
\end{align}
where $A\in\mathbb{Z}_{C+1}$ indexes the Hankel windows and $c\in\mathbb{Z}_{L}$ indexes delay coordinates within each window. Thus, the prefix produces exactly $C+1=\ell-L+1$ overlapping windows. We designate the final window as the \emph{query/evaluation} window,
\begin{align}
\alpha := C+1,
\qquad
R_{\alpha c X} := R_{(C+1)\,cX},
\end{align}
and treat the preceding $C$ windows as the \emph{context} set, $A\in\mathbb{Z}_{C}$. Each context window admits a one-step-ahead target contained within the prefix ($A=\mathbb{Z}_C$):
\begin{align}
y_{A X}
\;:=\;
R_{(A+1),\,L,\,X}
\;=\;
R_{(A+L)\,X}
\in\mathbb{R}^{D} \quad,
\end{align}
i.e., the target is the sample immediately following the end of the $A$-th delay window. By construction, the corresponding target for the query window $A=\alpha$ lies outside the observed prefix and is unavailable at adaptation time.

\subsection{Residuals \label{Residuals}}

Abstracting the IC-PCA motivation, we introduce a frozen encoder $\mathcal{E}$ and a frozen one-step forecasting decoder $\mathcal{D}$. The encoder maps each Hankel window to a latent coordinate,
\begin{align}
    R_{Ax} := \mathcal{E}(R_{A c X}), \qquad A\in\mathbb{Z}_{C+1},
\end{align}
and the decoder produces a global one-step prediction in ambient space,
\begin{align}
    Q_{AX} := \mathcal{D}(R_{Ax}), \qquad A\in\mathbb{Z}_{C+1}.
\end{align}
For the context windows $A=\mathbb{Z}_C$, the one-step-ahead target $y_{AX}$ is available within the prefix (see \S\ref{Hankelization}), and we define residuals:
\begin{align}
    \delta_{AX} := y_{AX} - Q_{AX}, \quad.
\end{align}
These residuals quantify the mismatch between the frozen global predictor and the locally observed dynamics in the current prefix, and they serve as supervision for the in-context correction evaluated at the query window $\alpha=C+1$.

\subsection{Latent kernel and context gating}

We define an RBF kernel over latent diffusion coordinates using parameters
$\beta>0$ (locality strength) and $\varepsilon>0$ (scale):
\begin{align}
k(R_{Ax},R_{A'x}) &:= \exp\!\left(-\beta\,\frac{\|R_{Ax}-R_{A'x}\|_2^2}{\varepsilon}\right).
\end{align}
We use the shorthand:
\begin{align}
k_{AA'} &:= k(R_{Ax},R_{A'x}), \\ 
k_{\alpha A} &:= k(R_{\alpha x},R_{Ax}),\\
k_{\alpha\alpha} &:= k(R_{\alpha x},R_{\alpha x}) = 1.
\end{align}
We optionally modulate the overall correction strength via a context-size gate
\begin{align}
\gamma_{\mathrm{ctx}} = \frac{C}{C+k_0}\in[0,1],
\end{align}
where $k_0>0$ sets the context scale at which corrections saturate.

\subsection{In-Context Gaussian-Process \label{ICGP}}

ICGP is an in--context adaptation layer built on top of the frozen NLSA encoder from \S\ref{NLSA} and the nonlinear GPLM decoder from \S\ref{GPLM}. The goal is to retain the global geometric regularity of diffusion coordinates while allowing rapid adaptation to a \emph{novel} time--series prefix without gradient-based retraining.

Conditioning the GP on the prefix residuals implements adaptation by locally re-centering the global predictor in latent space, using only context and without updating model parameters. The posterior mean residual correction at the query state is:
\begin{align}
    \delta_{\alpha X} = k_{\alpha A}\,\left({k}_{AA'}+\sigma^2 I_{AA'}\right)^{-1} \delta_{A'X} \in\mathbb{R}^{D},
    \label{lisa_gp_mean}
\end{align}
where repeated indices are summed. The scalar posterior variance (shared across $X\in\mathbb{Z}_D$ under the i.i.d.\ noise model) is:
\begin{align}
    s_\alpha^2 = k_{\alpha\alpha} - k_{\alpha A}\,\left({k}_{AA'}+\sigma^2 I_{AA'}\right)^{-1}k_{A'\alpha}.
    \label{lisa_gp_var}
\end{align}
A simple variance-based uncertainty gate is achieved with $\tau^2$, the variance scale parameter, in the confidence gate:
\begin{align}
    \gamma_{\mathrm{var}}(R_{\alpha x}) &= \frac{\tau^2}{\tau^2+s_\alpha^2}.
\end{align}
The ICGP one-step prediction in ambient space is:
\begin{align}
    R_{\alpha X} = Q_{\alpha X} + \gamma \gamma_{\mathrm{var}}\,\delta_{\alpha X}.
    \label{lisa_combined}
\end{align}
During autoregressive rollout, $R_{\alpha X}$ is appended to the current window to form the next query/eval window. Equation \eqref{lisa_combined} reduces to the baseline forecaster when $C=0$ (i.e. $\ell=L$).
Finally, ICGP can operate in a generative mode by sampling residuals from the GP posterior,
\begin{align}
    \delta^{\mathrm{sample}}_{\alpha X} &= \delta_{\alpha X} + \sqrt{s_\alpha^2+\sigma^2}\,\xi_X, \qquad \xi_X\sim\mathcal{N}(0,1),
\end{align}
which ``adds noise back'' in a calibrated way while preserving manifold-aware conditioning through the latent-space.

\subsection{In-Context Nadaraya-Watson \label{ICNW}}

Alternatively in--context adaptation layer that shares the same frozen encoder and decoder, but replaces the dense GP residual model with an \emph{attention-like, row-stochastic Markov operator} over context windows. That is the Nadaraya--Watson (kernel analog forecasting) estimate of the residual correction in diffusion-coordinate space. We begin just like ICGP, in \S\ref{Hankelization}, however the adaption stage is different. ICNW predicts the query residual by a kernel-weighted average of context residuals in diffusion-coordinate space. and normalize row-wise to obtain attention-like weights
\begin{align}
    k^+_{\alpha A}
    \coloneqq
    \frac{k_{\alpha A}}{\sum_{A'\in\mathbb{Z}_{C}} k_{\alpha A'}}, \qquad A\in\mathbb{Z}_{C}.
    \label{ICNW_weights}
\end{align}
By construction, $k^+_{\alpha A}\ge 0$ and $\sum_{A}k^+_{\alpha A}=1$, hence $k^+_\alpha=(k^+_{\alpha A})$ defines a row-stochastic Markov (attention-like) operator over the context set. The ICNW in-context residual estimate is the convex combination
\begin{align}
    \delta_{\alpha X}
    \coloneqq
    k^+_{\alpha A}\,\delta_{AX}\quad.
    \label{ICNW_residual_estimate}
\end{align}
Finally, the ICNW one-step prediction is then:
\begin{align}\label{ICNW_combined}
    R_{\alpha X} = Q_{\alpha X} + \gamma\,\delta_{\alpha X}\quad.
\end{align}
Equation \eqref{ICNW_combined} reduces to the baseline forecaster when $C=0$ (i.e. $\ell=L$). Given an initial prefix, we generate multi-step forecasts by iterating \eqref{ICNW_combined}: each prediction $R_{\alpha X}$ is appended to update the current delay window, the query diffusion coordinate $R_{\alpha x}$ is recomputed via the frozen encoder, and the attention-like weights \eqref{ICNW_weights} are updated at each step.

\subsection{LISA and ALSA}

We refer to the GPLM forecaster equipped with ICGP (\S\ref{ICGP}) as \textbf{LISA}. Replacing the GP correction with ICNW (\S\ref{ICNW}) yields an attention-like in-context mechanism; we refer to this variant as \textbf{ALSA}. In both cases, the encoder and baseline decoder are frozen after training, and adaptation occurs solely through the in-context residual correction computed from the test prefix.

\section{Experiments}
\label{sec:experiments}

This section describes the experimental protocol used to generate all plots and tables. We evaluate forecasting with and without in-context adaptation across three settings: (i) stationary chaotic attractors, (ii) forced/nonstationary dynamics via a controlled regime switch, and (iii) a real-world electricity load dataset. Interpretation of results and connections to related work are deferred to the Discussion.


\begin{figure}[t]
    \centering
    \includegraphics[width=1.0\linewidth]{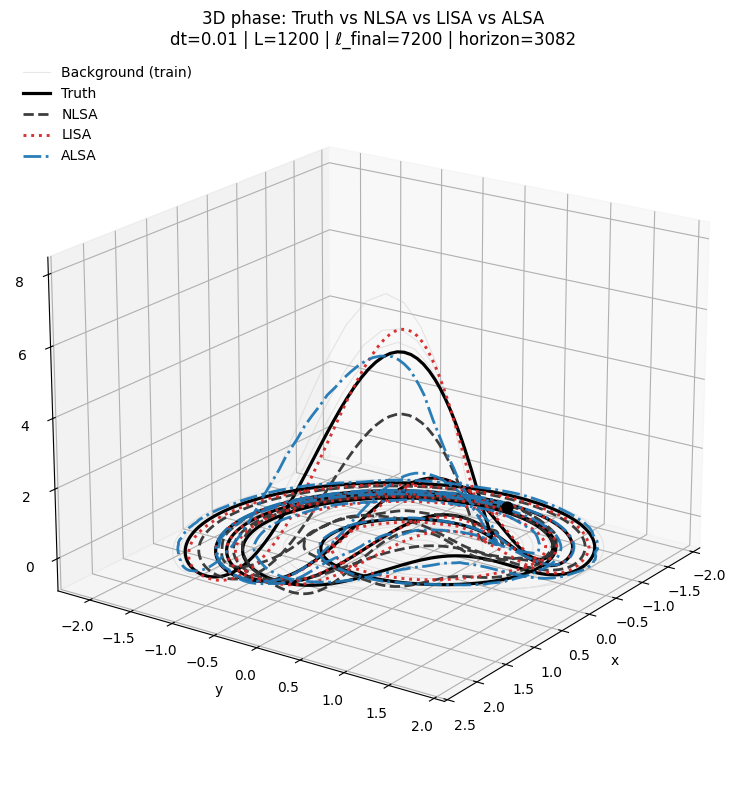}
    \caption{\textbf{Rössler: phase-space rollout visualization.} Gray shows a subsampled training trajectory (background). Black is a ground-truth test rollout from a fixed start. Colored curves are forecasts from the same start: the NLSA baseline (context $\ell=L$) and the in-context methods (LISA/ALSA with $\ell>L$). In chaotic systems, forecasts may dephase from the truth while remaining close to the attractor geometry; this visualization highlights attractor fidelity beyond strict pointwise tracking.}
    \label{fig:rossler_3d}
\end{figure}

\begin{figure}[t]
    \centering
    \includegraphics[width=1.0\linewidth]{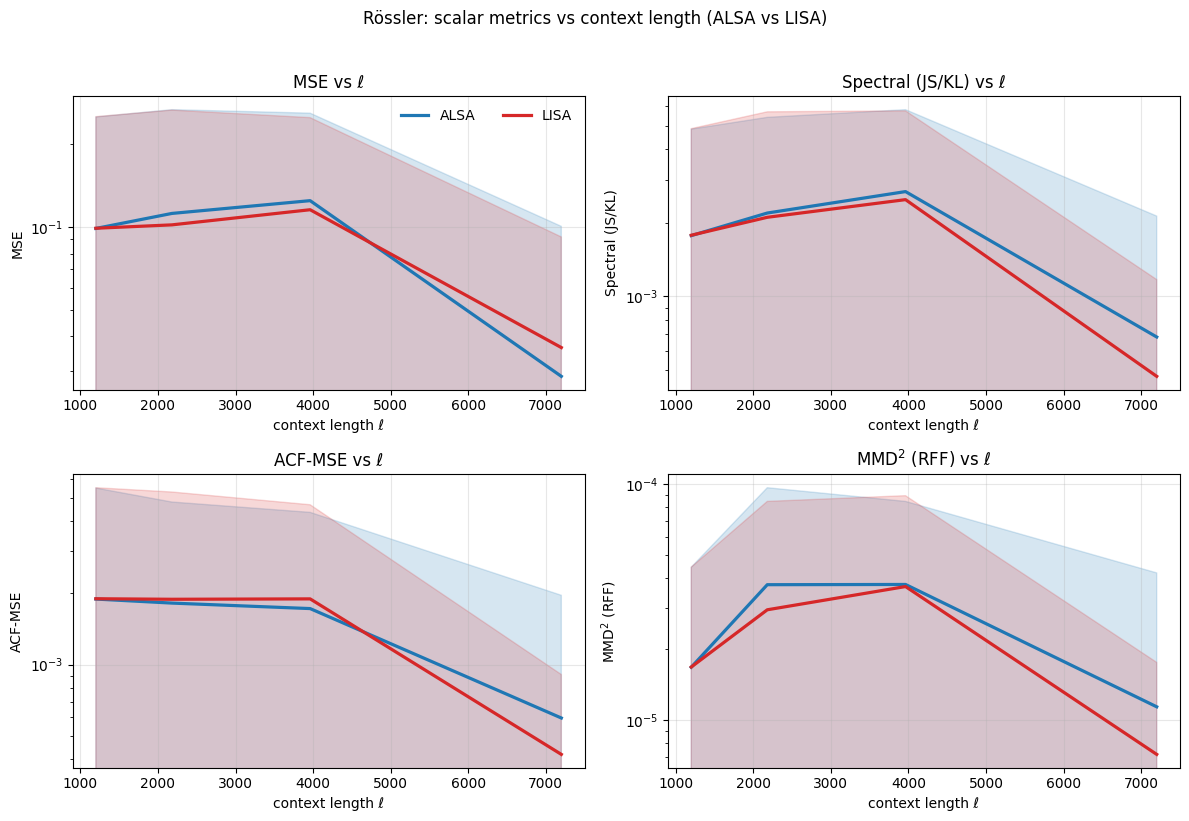}
    \caption{\textbf{Rössler: context-length response curves.} Scalar forecast metrics as a function of context length $\ell$ (multiples of the base window length $L$). Each curve reports mean $\pm$ one standard deviation across multiple test starts. Panels show MSE, spectral divergence computed from Welch PSD estimates (JS/KL-style), ACF-MSE, and MMD$^2$ (RFF).}
    \label{fig:rossler_metrics_vs_context}
\end{figure}

\subsection{Setup and evaluation protocol}
\label{sec:exp_setup}

Across experiments we compare \textbf{NLSA}, \cite{Giannakis2012_NLSA}, \textbf{LISA}, and \textbf{ALSA}, and (for the real-data experiment) a supervised neural baseline \textbf{PTST}, \cite{nie2023_patchtst64words}.
NLSA serves as a \emph{no in-context adaptation} baseline: it uses only the minimum history required by the base window length $L$, i.e.\ context $\ell=L$.
LISA and ALSA share the same base representation (fixed $L$ and latent rank) but additionally exploit longer prefixes $\ell > L$ at inference time via in-context adaptation. PTST is included only for the electricity dataset, where it provides a strong supervised reference.

Given a multivariate time series $R_{tX} \in \mathbb{R}^{N\times D}$, we select a forecast origin $t_0$ in the test segment and form a prefix (context) of length $\ell$ together with a horizon-$H$ target:
\begin{align*}
    \text{prefix } &= R_{tX}[t_0-\ell:t_0-1, :] = R_{iX}\quad, \\
    \text{truth }  &= R_{tX}[t_0:t_0+H-1,:] = R_{aX}\quad.
\end{align*}
Each method produces an autoregressive rollout. We either (i) sweep the context length $\ell$ to obtain context-response curves, or (ii) in the forced setting, fix a long context and sweep an in-context \emph{temperature} parameter that controls the locality of in-context matching (Section~\ref{sec:forced_attractor}).

For each dataset, models are trained on the same training segment. LISA and ALSA are trained \emph{once} for a fixed Takens/window length $L$ and latent rank (and other hyperparameters held fixed), so performance differences observed when varying $\ell$ reflect \emph{test-time use of additional context} rather than changes in training data or model capacity. PTST is trained once on the same training segment following its supervised setup.
To ensure comparisons reflect dynamical differences rather than scale, we standardize each coordinate using training statistics:
\begin{equation}
\tilde x(t)=\frac{x(t)-\mu_{\text{train}}}{s_{\text{train}}}.
\end{equation}
All models are trained and evaluated in standardized space. For stationary synthetic attractors, where trajectories are stationary after burn-in, we also report results under global standardization for convenience; for forced/nonstationary and real-data experiments we use train-only standardization to avoid leakage.

To reduce sensitivity to the choice of forecast origin, we evaluate using multiple start indices $\{t_0^{(n)}\}$ sampled from the admissible region of the test segment (ensuring $t_0 \ge \ell_{\max}$ and $t_0+H \le T_{\text{test}}$). Reported scalar metrics are aggregated across starts, and we plot mean $\pm$ one standard deviation when showing uncertainty bands.

Each rollout is compared to the ground truth using complementary metrics capturing pointwise accuracy and dynamical/statistical fidelity:
\begin{itemize}
    \item \textbf{MSE} and the horizon-resolved curve $\mathrm{MSE}(h)$ to characterize error growth with forecast step;
    \item \textbf{ACF-MSE}, the mean-squared discrepancy between normalized autocorrelation functions, to compare temporal dependence structure;
    \item \textbf{Spectral divergence} computed from Welch power spectral density estimates (reported as a JS/KL-style divergence), to compare frequency content in a phase-tolerant manner;
    \item \textbf{MMD$^2$} between state-sample distributions (approximated using random Fourier features), to assess distributional similarity / attractor (``climate'') fidelity.
\end{itemize}
These metrics separate strict trajectory tracking from longer-run statistical agreement, which is essential in chaotic settings where dephasing is expected. The metric definitions are summarized in Table~\ref{tab:error-metrics}.

\subsection{Stationary chaotic attractors \label{sec:stationary_attractor}}

We first evaluate on stationary dynamical systems whose trajectories lie on a fixed attractor. The goal is to verify that in-context adaptation does not degrade a stable baseline and can improve forecasting when additional recent history is informative. We generate long trajectories from canonical chaotic systems (e.g.\ R\"ossler and others), discard an initial burn-in so the system settles onto the attractor, and split the remaining trajectory into disjoint training and test segments. Models are trained on the training segment only. The set of attractors considered is summarized in Table~\ref{tab:attractors_lyapunov_reference}.

For a fixed forecast horizon $H$, we sweep the context length from the minimum $\ell=L$ (the NLSA baseline regime) up to several multiples of $L$. For each $\ell$ we run the multi-start evaluation on the test segment and compute the metrics in Section~\ref{sec:exp_setup}. We report:
(i) scalar metric-versus-$\ell$ curves with uncertainty bands (Fig.~\ref{fig:rossler_metrics_vs_context}), and
(ii) for selected context lengths, the mean MSE-by-horizon curve $\mathrm{MSE}(h)$ to visualize how quickly errors grow with forecast step.
For 3D attractors we additionally visualize representative rollouts in phase space (Fig.~\ref{fig:rossler_3d}) to illustrate attractor fidelity even under dephasing.

\subsection{Forced (nonstationary) attractor \label{sec:forced_attractor}}

To test adaptation under distribution shift, we construct a regime-switch Lorenz--63 experiment in which the dynamical parameters change over time. We train on a trajectory segment generated under \emph{regime A} and evaluate on a subsequent segment generated under a distinct \emph{regime B}. This yields a clean ``train on A $\rightarrow$ test on B'' setup where generalization requires adaptation rather than interpolation.

We simulate the Lorenz--63 system with step size $\Delta t=0.01$ and discard an initial burn-in segment to remove transients. We then construct a regime-switch split aligned with a single parameter change: the training segment contains only regime $A$, using standard chaotic parameters $(\sigma=10,\rho=28,\beta=8/3)$, and the test segment contains only regime $B$. A shifted configuration (e.g., $\sigma=16,\rho=50,\beta=3$), inducing a systematic change in the attractor geometry and local temporal statistics.

We evaluate two variations in regime B:
(i) a \emph{context-length sweep}, which measures how additional context improves adaptation and long-horizon stability, reporting the same scalar metrics and MSE-by-horizon curves as in the stationary case; and
(ii) a \emph{temperature sweep} at fixed long context (e.g. $\ell=16L$), which scales the effective similarity lengthscale used for in-context matching (larger temperature yields smoother/more global mixing; smaller temperature yields sharper/more local matching).
We visualize representative predictions using grids of 3D phase plots and quantify performance using the same metrics as above.

\begin{figure}[t]
    \centering
    \includegraphics[width=1.0\linewidth]{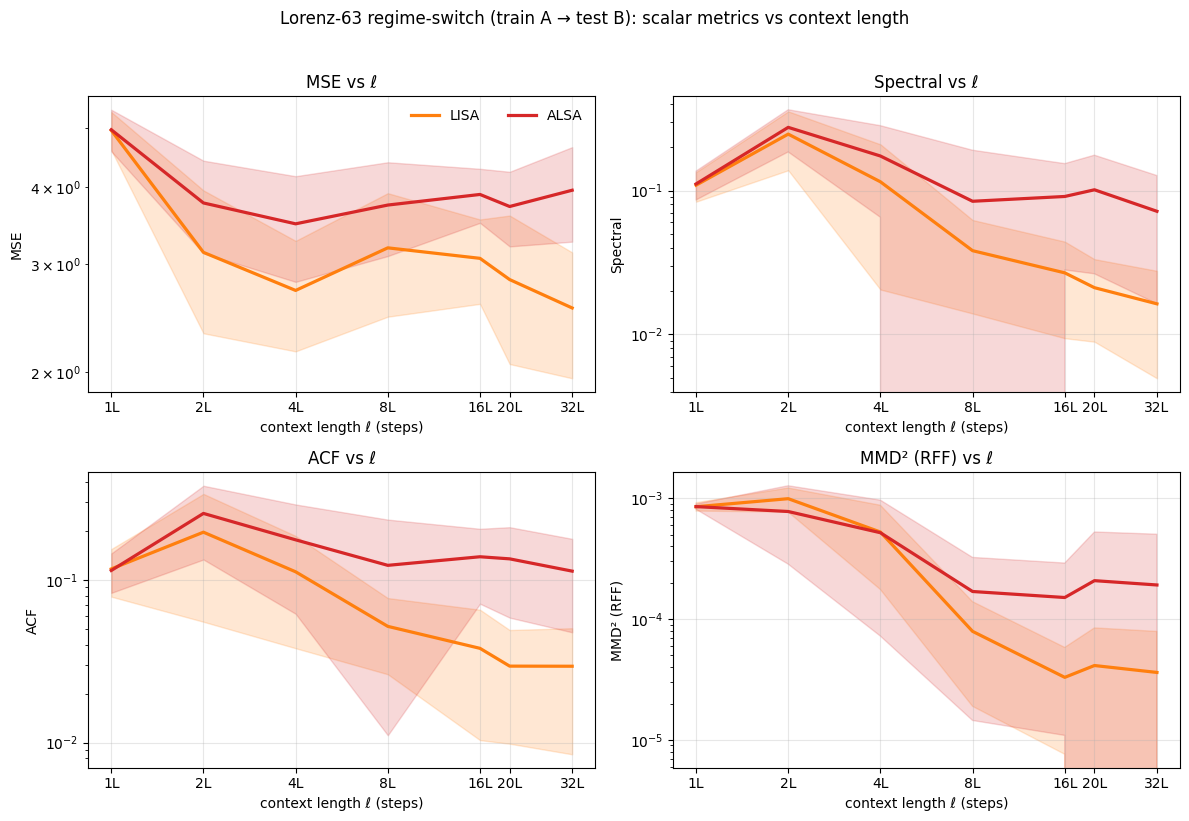}
    \caption{\textbf{Regime-switch Lorenz--63: scalar metrics vs context length in regime B.} Forecast metrics in the shifted regime B as a function of context length $\ell$, aggregated as mean $\pm 1$ standard deviation over multiple test starts.}
    \label{fig:forcing_context_mse}
\end{figure}

\begin{figure}[t]
    \centering
    \includegraphics[width=1.0\linewidth]{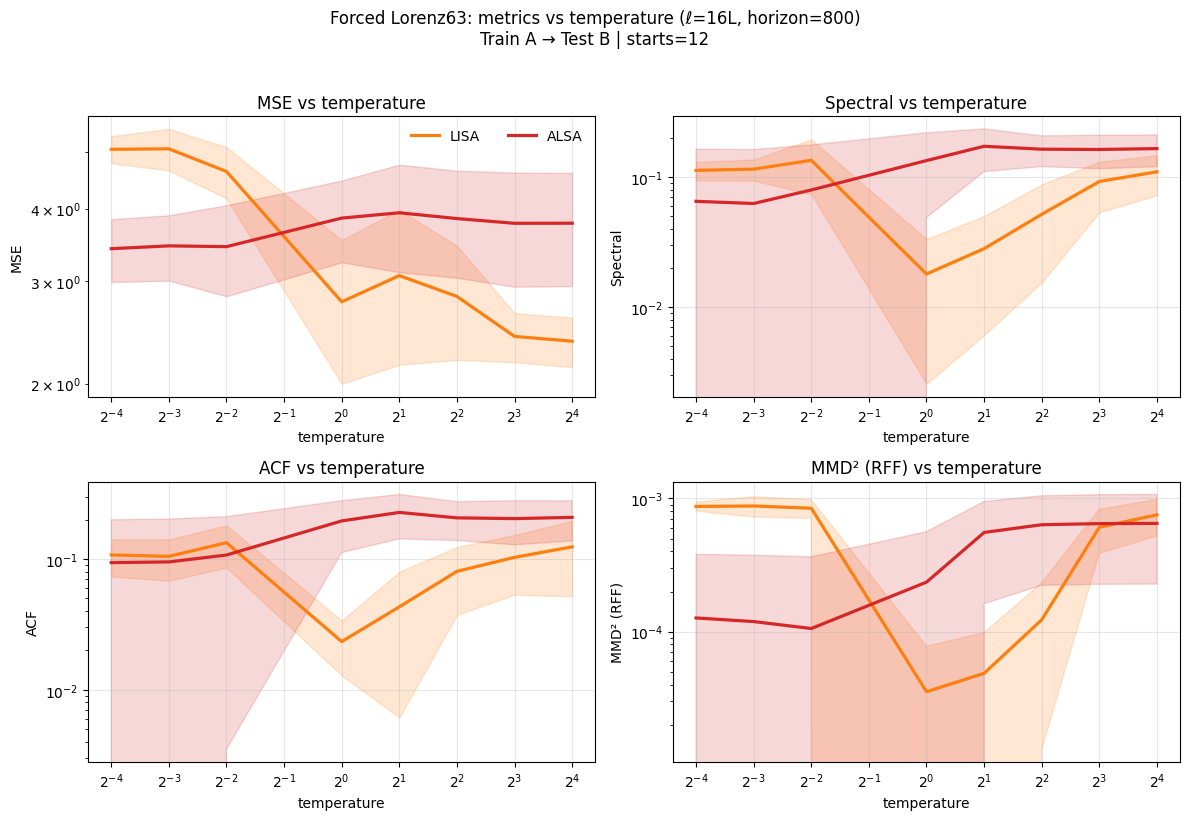}
    \caption{\textbf{Regime-switch Lorenz--63: scalar metrics vs in-context temperature at fixed long context.} Metrics in regime B as a function of a temperature parameter that controls the locality of in-context matching (larger values correspond to smoother/more global mixing).}
    \label{fig:forcing_temp_mse}
\end{figure}

\begin{figure*}[t]
    \centering
    \includegraphics[width=1.0\linewidth]{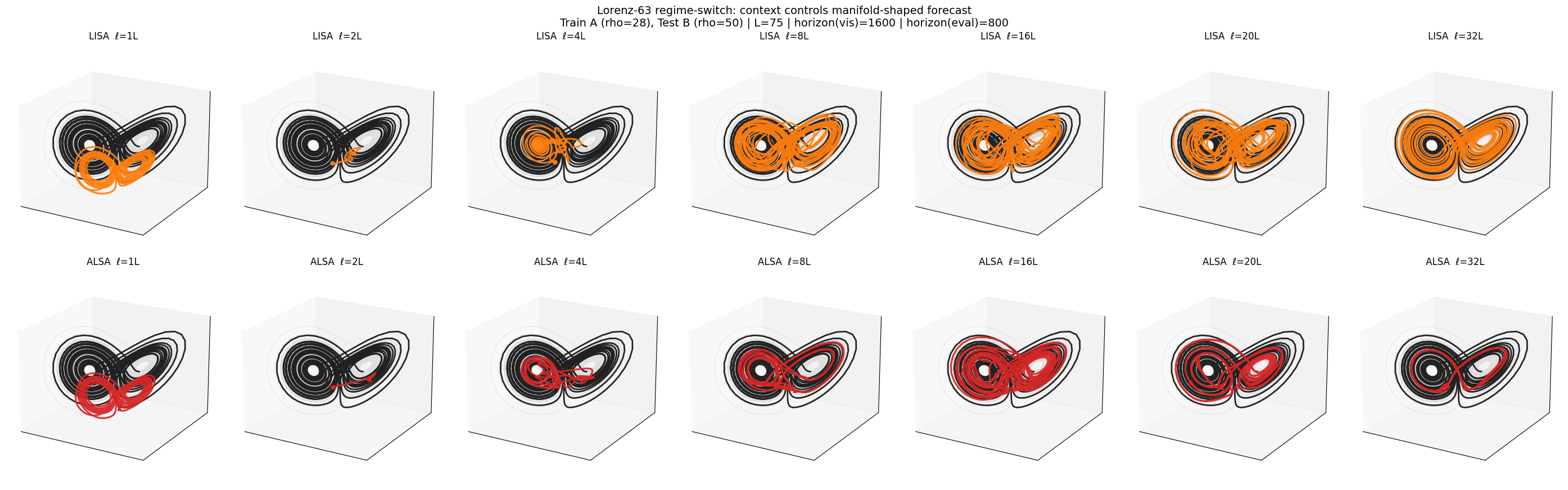}
    \caption{\textbf{Regime-switch Lorenz--63: 3D context grid in regime B.} 3D phase portraits for a fixed test start in regime B, sweeping context length $\ell$ (multiples of $L$). Each panel overlays the model forecast (colored) on a background of the regime-B trajectory (gray), highlighting how longer context changes the predicted manifold geometry and long-horizon stability.}
    \label{fig:forcing_context}
\end{figure*}

\begin{figure*}[t]
    \centering
    \includegraphics[width=1.0\linewidth]{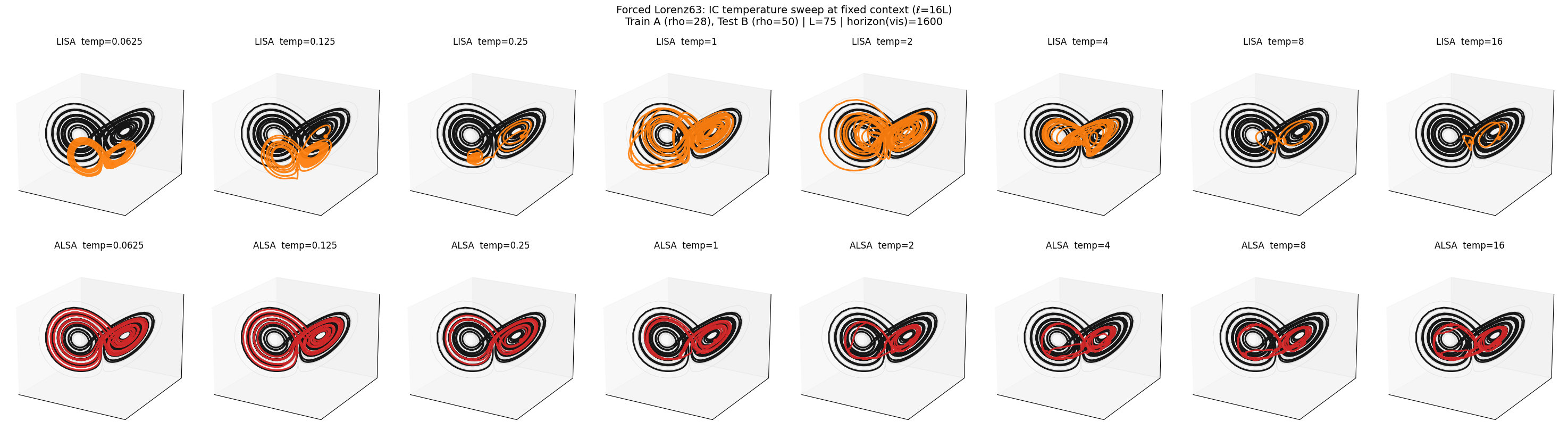}
    \caption{\textbf{Regime-switch Lorenz--63: 3D temperature grid at fixed long context.} 3D phase portraits for a fixed test start in regime B, sweeping the in-context temperature parameter while holding $\ell$ fixed (e.g. $\ell=16L$). Larger temperatures yield smoother, more global in-context mixing; smaller temperatures yield sharper, more local matching.}
    \label{fig:forcing_temp}
\end{figure*}

\subsection{Real-world electricity load forecasting \label{sec:real_data}}

Finally, we evaluate on a real-world, high-dimensional time series: county-level electricity load, \cite{OEDI_Dataset_8562}. The data are resampled hourly, transformed with $\log(1+x)$ to stabilize variance, and standardized using training statistics only. We use a chronological split (early segment for training, later segment for testing) to reflect realistic deployment.

We sweep context length over a small set of multiples of $L$ (e.g. $\ell \in \{L,2L,4L,10L\}$) and evaluate with the same multi-start protocol and metrics as in the synthetic experiments. In addition to NLSA, LISA, and ALSA, we include PTST as a supervised neural baseline. We report:
(i) scalar metric-versus-$\ell$ curves,
(ii) MSE-by-horizon curves highlighting short- versus long-horizon behavior, and
(iii) representative rollouts for a small subset of features in original (unstandardized) units.

\begin{figure}[t]
    \centering
    \includegraphics[width=1.0\linewidth]{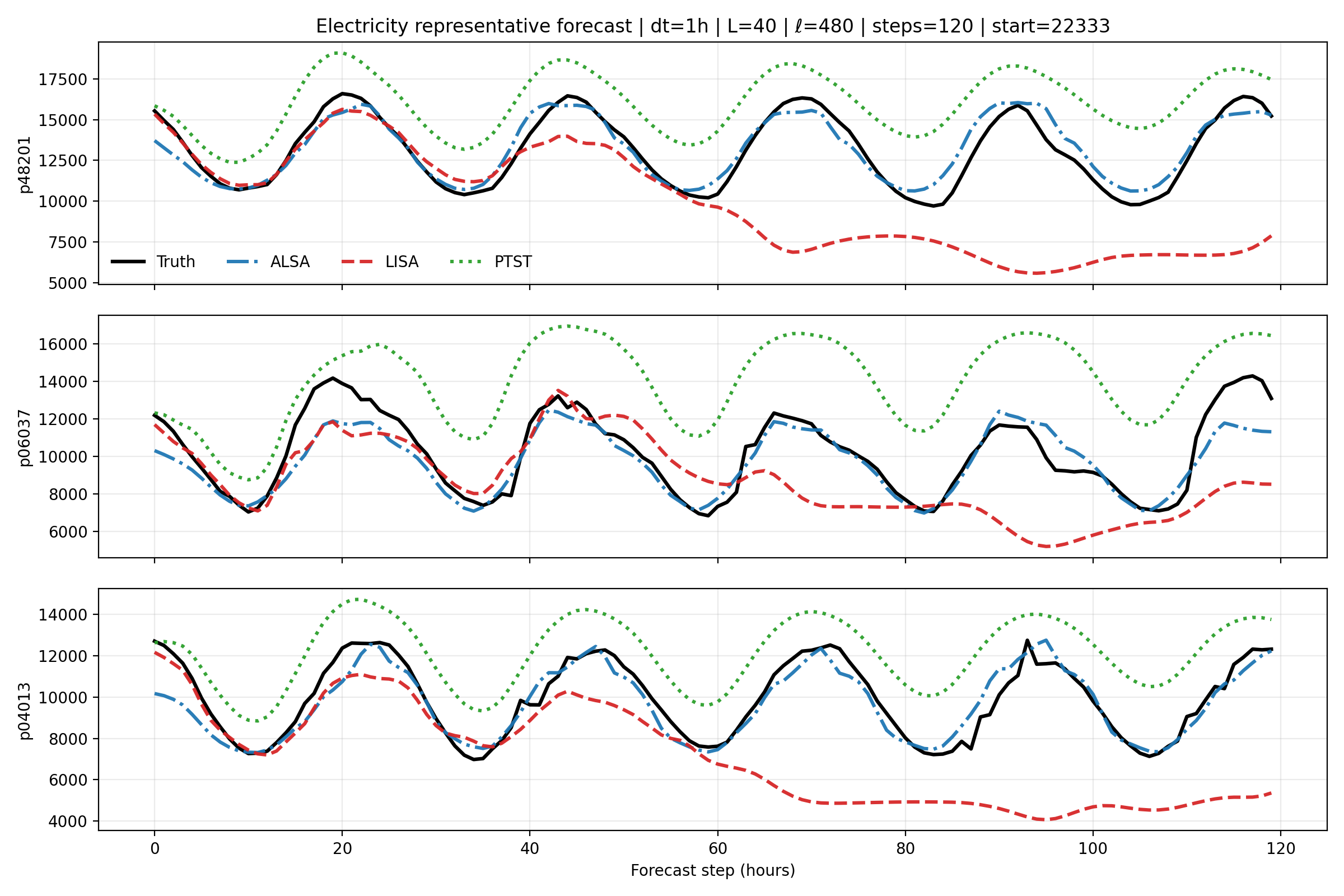}
    \caption{\textbf{Electricity load forecasting: representative forecasts in original units.} Example rollouts for three selected counties/features from a single test start. We plot the ground truth and model predictions after reversing standardization and the $\log(1+x)$ transform, illustrating qualitative behavior at a long context (e.g.\ $\ell=10L$).}
    \label{fig:electricity_display}
\end{figure}

\begin{figure}[hbt!]
    \centering
    \includegraphics[width=1.0\linewidth]{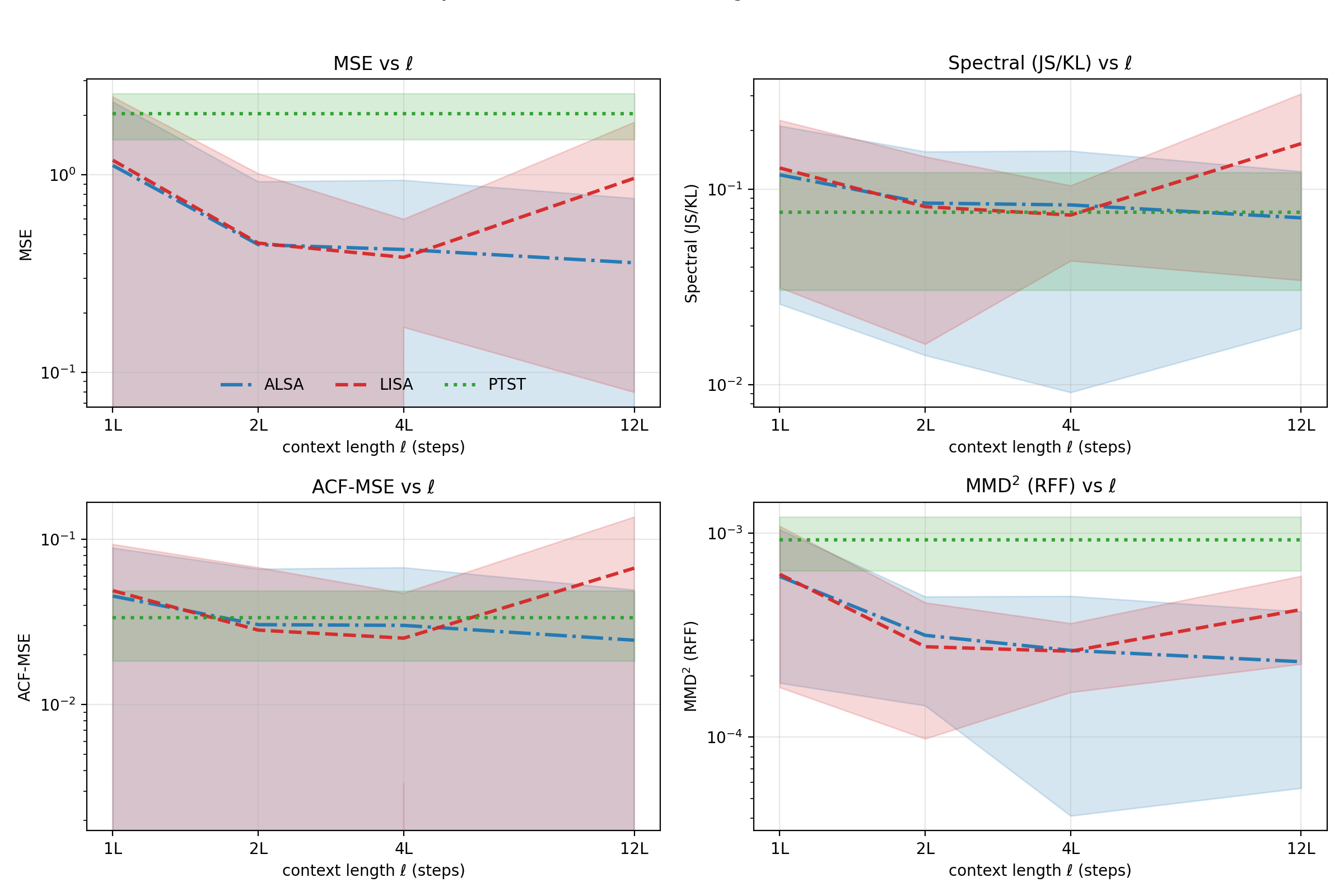}
    \caption{\textbf{Electricity load forecasting: scalar metrics vs context length.} For each context length $\ell$ (multiples of $L$), we report mean $\pm 1$ standard deviation over multiple forecast starts on the test period. Panels show MSE, spectral divergence from Welch PSD (JS/KL-style), ACF-MSE, and MMD$^2$ (RFF), comparing LISA, ALSA, and PTST.}
    \label{fig:electricity_mse}
\end{figure}

\begin{figure}[t]
    \centering
    \includegraphics[width=1.0\linewidth]{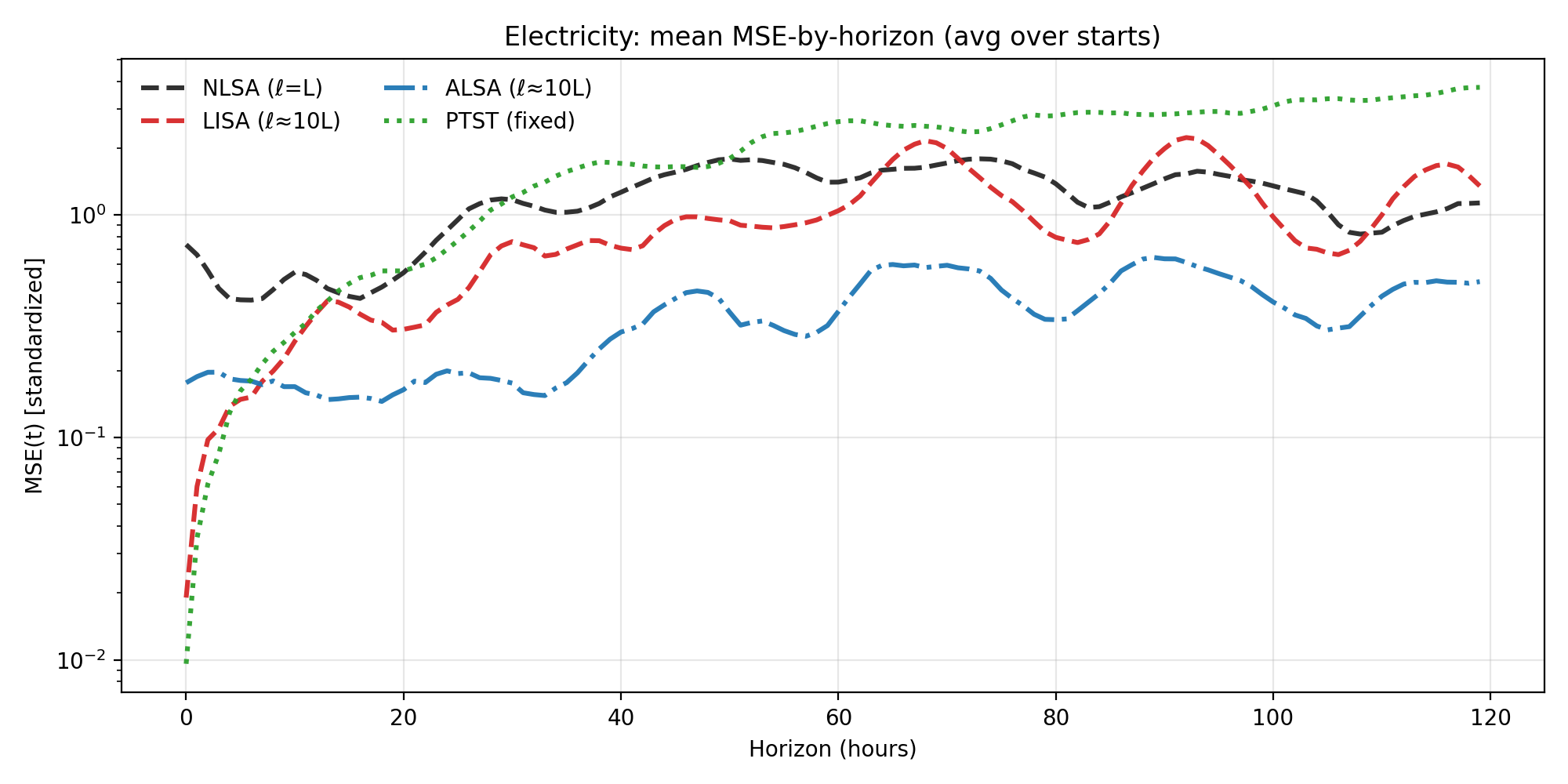}
    \caption{\textbf{Electricity load forecasting: error growth with horizon.} Mean per-step error curve $\mathrm{MSE}(h)$ (averaged over features and forecast starts) comparing the NLSA baseline ($\ell=L$) to long-context forecasts (e.g.\ LISA/ALSA at $\ell\approx 10L$). PTST (fixed window) is shown when available.}
    \label{fig:electricity_mse_t}
\end{figure}

\section{Discussion}

Across all experiments we report both qualitative \emph{dynamical fidelity} and quantitative error metrics. Phase-portrait visualizations (Figs.~\ref{fig:rossler_3d}, \ref{fig:forcing_context}, \ref{fig:forcing_temp}, \ref{fig:electricity_display}) provide a rapid check that forecasts remain on (or close to) the correct invariant set, while metric panels (Figs.~\ref{fig:rossler_metrics_vs_context}, \ref{fig:forcing_context_mse}, \ref{fig:forcing_temp_mse}, \ref{fig:electricity_mse}, \ref{fig:electricity_mse_t}) quantify point-wise and statistical discrepancies. Visual confirmation is particularly important for chaotic systems: due to sensitive dependence on initial conditions, trajectories inevitably dephase even when a method captures the correct attractor geometry. In this regime, ``looking right'' in phase space is evidence of dynamical consistency that complements horizon-wise errors.

In the stationary Rössler experiment (Fig.~\ref{fig:rossler_metrics_vs_context}), we observe a non-monotonic dependence on context length: errors worsen at intermediate $\ell$ before improving again at larger $\ell$. This suggests that moderate context can introduce misleading analogs (or unstable in-context updates), whereas sufficiently long context provides enough recurrence structure for the in-context mechanism to stabilize and improve both short-horizon accuracy and long-horizon fidelity. Notably, this mid-context degradation is less pronounced in the nonstationary regime-switch setting and on electricity data (Figs.~\ref{fig:forcing_context_mse}, \ref{fig:electricity_mse}), where additional context more consistently improves performance.

In the regime-switch Lorenz--63 experiment, both LISA and ALSA adapt to an unseen test-time attractor using \emph{in-context information only} (Figs.~\ref{fig:forcing_context}, \ref{fig:forcing_context_mse}). Increasing $\ell$ yields progressively better geometric agreement in phase space, indicating that longer prefixes supply more useful analog information for adaptation. Performance peaks at an intermediate-to-large context (here around $\ell=16L$ for ALSA), after which gains saturate. We further probe the role of in-context selectivity via a ``temperature'' sweep at fixed $\ell$ (Figs.~\ref{fig:forcing_temp}, \ref{fig:forcing_temp_mse}). At very low temperature (highly local updates), LISA fails to adapt reliably, while ALSA remains stable and can even improve; at moderate temperature, LISA achieves its best performance; and at high temperature (approaching a globally mixed / linearized limit) both methods degrade, consistent with over-smoothing the local geometry needed for accurate analog selection.

Finally, on the real electricity dataset (with noise, seasonal structure, and potential weak nonstationarity), we find that total forecast error generally decreases with larger context for both ALSA and LISA (Fig.~\ref{fig:electricity_mse}). However, LISA can degrade at the longest contexts, consistent with sensitivity to noisy or weakly relevant history, while ALSA continues to improve. PTST exhibits lower error at short horizons but substantially larger long-horizon error, leading to worse total MSE over the full 120-step rollout (Fig.~\ref{fig:electricity_mse_t}). A representative qualitative forecast for three counties is shown in Fig.~\ref{fig:electricity_display}.

\subsection{Related Work}

Nonlinear forecasting for chaos often relies on local recurrence in reconstructed state space, dating back to classic results on predicting chaotic time series in the presence of noise \cite{sugihara1990_nonlinearForecastingChaosNoise}. Kernel analog forecasting extends these ideas by combining similarity kernels with operator-theoretic structure and has been shown to pair naturally with NLSA-style spectral representations \cite{zhao2016_analogForecastingDAkernels, giannakis2021_streamingKAF}. Chaotic attractors remain a standard testbed—both for benchmarking \cite{gilpin2021_chaosBenchmarkForecasting} and for demonstrations in alternative computing substrates such as reservoir/quantum-inspired models on Lorenz–63 \cite{steinegger2025_chaos3dQRC4qubit}. Separately, time-series foundation models \cite{ansari2024_chronosTimeSeries} and recent work on in-context prediction for time series \cite{lu2024_inContextTSP} motivate studying test-time adaptation mechanisms. We build on these directions by quantifying how in-context mechanisms affect attractor fidelity and nonstationary adaptation across synthetic dynamical systems and real electricity load data.

\section{Conclusion}

We introduced \emph{Laplacian In-context Spectral Analysis} (LISA), a framework that augments NLSA-style geometric time-series models with a lightweight \emph{in-context mechanism} (ICM). Across three settings---stationary chaotic attractors, a regime-switch (nonstationary) dynamical system, and real hourly electricity demand---we found that in-context adaptation can improve forecasting performance and dynamical fidelity when additional history is informative. In the stationary setting, the ICM does not destabilize the base model and can improve long-horizon attractor consistency even when pointwise trajectories dephase. In the regime-switch setting, longer context enables prompt-time adaptation to an unseen attractor without parameter updates, and a temperature sweep highlights the importance of tuning the selectivity of the in-context update. On electricity data, longer context typically reduces aggregate error for the in-context methods, while highlighting a trade-off between short-horizon accuracy and long-horizon stability across model classes.

A primary limitation of our approach is the dependence on the Takens delay-window length $L$ (and related embedding/feature choices). In practice, $L$ is data-dependent and there may exist an ``ideal'' range that best resolves the underlying state; performance and the benefits of adaptation can vary outside this range. Additional limitations include sensitivity to observation noise or weakly relevant history at long contexts, and computational costs that grow with context length if not carefully subsampled or approximated. Addressing these issues suggests several directions for future work: automated or multi-scale selection of $L$, context-selection strategies that downweight irrelevant history, and theoretical characterization of when and why in-context adaptation helps (e.g., in terms of recurrence statistics or effective Lyapunov time).

\subsection{Contributions}
\begin{itemize}
    \item A geometric ICL-style forecasting framework. We propose LISA, which couples NLSA diffusion-coordinate representations with an explicit decoder (GPLM) and a prompt-time in-context update mechanism.
    \item Two in-context mechanisms for continuous sequences. We develop and study two lightweight adapters---a Markov-operator update and a GP-based update---that modify inference-time behavior without changing global parameters.
    \item A multi-regime evaluation protocol. We evaluate stationary chaotic forecasting (attractor fidelity under dephasing), nonstationary regime-switch adaptation (prompt-only adaptation), and noisy real-world electricity load forecasting.
\end{itemize}

\newpage
\bibliographystyle{icml2026}
\bibliography{example_paper}

@inproceedings{Takens1981,
  title={Detecting Strange Attractors in Turbulence},
  author={Takens, Floris},
  booktitle={\href{https://link.springer.com/chapter/10.1007/BFb0091924}{Dynamical Systems and Turbulence, Warwick 1980}},
  pages={366--381},
  year={1981},
  publisher={Springer-Verlag}
}

@article{candanedo2024_DMAE,
  author  = {Candanedo, Julio},
  title   = {Diffusion Map Autoencoder},
  journal = {\href{https://arxiv.org/abs/2409.05901}{arXiv:2409.05901}},
  year    = {2024},
}

@inproceedings{nie2023_patchtst64words,
  author    = {Nie, Yuqi and Nguyen, Nam H. and Sinthong, Phanwadee and Kalagnanam, Jayant},
  title     = {A Time Series is Worth 64 Words: Long-term Forecasting with Transformers},
  booktitle = {\href{https://openreview.net/group?id=ICLR.cc/2023/Conference}{ICLR}},
  year      = {2023},
  doi       = {10.48550/arXiv.2211.14730},
  note      = {\href{https://arxiv.org/abs/2211.14730}{arXiv:2211.14730}}
}

@inproceedings{vaswani2017attention,
  title     = {Attention Is All You Need},
  author    = {Vaswani, Ashish and Shazeer, Noam and Parmar, Niki and Uszkoreit, Jakob and Jones, Llion and Gomez, Aidan N. and Kaiser, Lukasz and Polosukhin, Illia},
  booktitle = {\href{https://papers.neurips.cc/paper/7181-attention-is-all-you-need.pdf}{NeurIPS}},
  year      = {2017},
  note      = {\href{https://arxiv.org/abs/1706.03762}{arXiv:1706.03762}}
}

@article{steinegger2025_chaos3dQRC4qubit,
  author       = {Steinegger, Joel and Räth, Christoph},
  title        = {Predicting three-dimensional chaotic systems with four qubit quantum systems},
  journal      = {\href{https://doi.org/10.1038/s41598-025-87768-0}{Sci. Rep.}},
  year         = {2025},
  volume       = {15},
  pages        = {6201},
  doi          = {10.1038/s41598-025-87768-0},
  note         = {\href{https://arxiv.org/abs/2501.15191}{arXiv:2501.15191}},
  eprint       = {2501.15191},
  archivePrefix= {arXiv},
  primaryClass = {quant-ph},
}

@article{zhao2016_analogForecastingDAkernels,
  author        = {Zhao, Zhizhen and Giannakis, Dimitrios},
  title         = {Analog forecasting with dynamics-adapted kernels},
  journal       = {\href{https://doi.org/10.1088/0951-7715/29/9/2888}{Nonlinearity}},
  year          = {2016},
  volume        = {29},
  number        = {9},
  pages         = {2888--2939},
  doi           = {10.1088/0951-7715/29/9/2888},
  note          = {\href{https://arxiv.org/abs/1412.3831}{arXiv:1412.3831}},
  eprint        = {1412.3831},
  archivePrefix = {arXiv},
  primaryClass  = {physics.data-an},
}

@misc{OEDI_Dataset_8562,
  title        = {Hourly Electricity Demand Profiles for Each County in the Contiguous United States},
  author       = {Obika, Kodi and Cole, Wesley and Rivers, Marie},
  year         = {2025},
  howpublished = {\href{https://data.openei.org/submissions/8562}{Open Energy Data Initiative (OEDI), National Renewable Energy Laboratory}},
  note         = {Accessed: 2026-01-28}
}

@article{Giannakis2012_NLSA,
  title={Nonlinear Laplacian Spectral Analysis for Time Series with Intermittency and Low-Frequency Variability},
  author={Giannakis, Dimitrios and Majda, Andrew J},
  journal={\href{https://www.pnas.org/doi/full/10.1073/pnas.1118984109}{PNAS}},
  volume={109},
  number={7},
  pages={2222--2227},
  year={2012},
  publisher={National Acad Sciences}
}

@book{Rasmussen2005_GP,
  title={\href{https://direct.mit.edu/books/oa-monograph/2320/Gaussian-Processes-for-Machine-Learning}{Gaussian Processes for Machine Learning}},
  author={Williams, Christopher and Rasmussen, Carl},
  publisher={The MIT Press},
  year={2005}
}

@article{Lorenz1963,
  title={Deterministic Nonperiodic Flow},
  author={Lorenz, Edward N.},
  year={1963},
  journal={\href{https://journals.ametsoc.org/view/journals/atsc/20/2/1520-0469_1963_020_0130_dnf_2_0_co_2.xml}{JAS}},
}

@article{Belkin2003,
  title={Laplacian Eigenmaps for Dimensionality Reduction and Data Representation},
  author={Belkin, Mikhail and Niyogi, Partha},
  year={2003},
  journal={\href{https://ieeexplore.ieee.org/document/6789755}{Neural Computation}}
}

@inproceedings{lawrence2004_gplvm,
  author    = {Lawrence, Neil D.},
  title     = {Gaussian Process Latent Variable Models for Visualisation of High Dimensional Data},
  booktitle = {\href{https://proceedings.neurips.cc/paper/2003}{NeurIPS}},
  series    = {Adv. Neural Inf. Process. Syst.},
  volume    = {16},
  pages     = {329--336},
  year      = {2004},
  editor    = {Thrun, Sebastian and Saul, Lawrence K. and Schölkopf, Bernhard},
  publisher = {MIT Press},
}

@article{sugihara1990_nonlinearForecastingChaosNoise,
  author  = {Sugihara, George and May, Robert M.},
  title   = {Nonlinear forecasting as a way of distinguishing chaos from measurement error in time series},
  journal = {\href{https://doi.org/10.1038/344734a0}{Nature}},
  year    = {1990},
  volume  = {344},
  pages   = {734--741},
  doi     = {10.1038/344734a0},
}

@inproceedings{gilpin2021_chaosBenchmarkForecasting,
  author    = {Gilpin, William},
  title     = {Chaos as an interpretable benchmark for forecasting and data-driven modelling},
  booktitle = {\href{https://datasets-benchmarks-proceedings.neurips.cc/paper_files/paper/2021/hash/ec5decca5ed3d6b8079e2e7e7bacc9f2-Abstract-round2.html}{NeurIPS Datasets \& Benchmarks}},
  year      = {2021},
  editor    = {Vanschoren, Joaquin and Yeung, Sai-Kit},
  note      = {\href{https://arxiv.org/abs/2110.05266}{arXiv:2110.05266}},
}

@misc{lorenz_predictability_ecmwf_1995,
  author    = {E. N. Lorenz},
  title     = {Predictability: a problem partly solved},
  year      = {1995},
  howpublished = {\href{https://www.ecmwf.int/en/elibrary/75462-predictability-problem-partly-solved}{Seminar on Predictability, 4--8 September 1995, Vol. I}},
  pages     = {1--18},
  publisher = {ECMWF},
  address   = {Reading, UK},
}

@article{ansari2024_chronosTimeSeries,
  author        = {Ansari, Abdul Fatir and Stella, Lorenzo and Turkmen, Caner and Zhang, Xiyuan and Mercado, Pedro and Shen, Huibin and Shchur, Oleksandr and Rangapuram, Syama Sundar and Pineda Arango, Sebastian and Kapoor, Shubham and Zschiegner, Jasper and Maddix, Danielle C. and Wang, Hao and Mahoney, Michael W. and Torkkola, Kari and Wilson, Andrew Gordon and Bohlke-Schneider, Michael and Wang, Yuyang},
  title         = {Chronos: Learning the Language of Time Series},
  journal       = {\href{https://openreview.net/forum?id=gerNCVqqtR}{TMLR}},
  year          = {2024},
  month         = nov,
  doi           = {10.48550/arXiv.2403.07815},
  note          = {\href{https://arxiv.org/abs/2403.07815}{arXiv:2403.07815}},
  eprint        = {2403.07815},
  archivePrefix = {arXiv},
  primaryClass  = {cs.LG},
}

@article{lu2024_inContextTSP,
  author        = {Lu, Jiecheng and Sun, Yan and Yang, Shihao},
  title         = {In-context Time Series Predictor},
  year          = {2024},
  doi           = {10.48550/arXiv.2405.14982},
  journal       = {\href{https://arxiv.org/abs/2405.14982}{arXiv:2405.14982}},
  eprint        = {2405.14982},
}

@article{rossler1976_continuousChaosEq,
  author  = {Rössler, Otto E.},
  title   = {An equation for continuous chaos},
  journal = {\href{https://doi.org/10.1016/0375-9601(76)90101-8}{Phys. Lett. A}},
  year    = {1976},
  volume  = {57},
  number  = {5},
  pages   = {397--398},
  month   = jul,
  doi     = {10.1016/0375-9601(76)90101-8},
}

@article{holmes1979_strangeAttractorOscillator,
  author  = {Holmes, Philip},
  title   = {A nonlinear oscillator with a strange attractor},
  journal = {\href{https://doi.org/10.1098/rsta.1979.0068}{Phil. Trans. R. Soc. Lond. A}},
  year    = {1979},
  volume  = {292},
  number  = {1394},
  pages   = {419--448},
  month   = oct,
  doi     = {10.1098/rsta.1979.0068},
}

@book{sprott2003_chaosTimeSeries,
  author    = {Sprott, Julien Clinton},
  title     = {Chaos and Time-Series Analysis},
  publisher = {\href{https://doi.org/10.1093/oso/9780198508397.001.0001}{Oxford Scholarship Online}{Oxford Univ. Press}},
  address   = {Oxford},
  year      = {2003},
  isbn      = {9780198508397},
  doi       = {10.1093/oso/9780198508397.001.0001},
}

@article{matsumoto1985_doubleScroll,
  author  = {Matsumoto, Takashi and Chua, Leon O. and Komuro, Motomasa},
  title   = {The double scroll},
  journal = {\href{https://doi.org/10.1109/TCS.1985.1085791}{IEEE Trans. Circuits Syst.}},
  year    = {1985},
  volume  = {32},
  number  = {8},
  pages   = {797--818},
  month   = aug,
  doi     = {10.1109/TCS.1985.1085791},
}

@book{strogatz2018_nonlinearDynamicsChaos,
  author    = {Strogatz, Steven H.},
  title     = {Nonlinear Dynamics and Chaos: With Applications to Physics, Biology, Chemistry, and Engineering},
  edition   = {2},
  publisher = {\href{https://doi.org/10.1201/9780429492563}{DOI:10.1201/9780429492563}{CRC Press}},
  address   = {Boca Raton, FL},
  year      = {2018},
  doi       = {10.1201/9780429492563},
  isbn      = {9780429492563},
}

@article{coifman2006_diffusionMaps,
  author  = {Coifman, Ronald R. and Lafon, Stéphane},
  title   = {Diffusion maps},
  journal = {\href{https://doi.org/10.1016/j.acha.2006.04.006}{ACHA}},
  year    = {2006},
  volume  = {21},
  number  = {1},
  pages   = {5--30},
  doi     = {10.1016/j.acha.2006.04.006},
}

@inproceedings{brown2020_fewShotLearners,
  author    = {Brown, Tom B. and Mann, Benjamin and Ryder, Nick and Subbiah, Melanie and Kaplan, Jared D. and Dhariwal, Prafulla and Neelakantan, Arvind and Shyam, Pranav and Sastry, Girish and Askell, Amanda and Agarwal, Sandhini and Herbert-Voss, Ariel and Krueger, Gretchen and Henighan, Tom and Child, Rewon and Ramesh, Aditya and Ziegler, Daniel M. and Wu, Jeffrey and Winter, Clemens and Hesse, Christopher and Chen, Mark and Sigler, Eric and Litwin, Mateusz and Gray, Scott and Chess, Benjamin and Clark, Jack and Berner, Christopher and McCandlish, Sam and Radford, Alec and Sutskever, Ilya and Amodei, Dario},
  title     = {Language Models are Few-Shot Learners},
  booktitle = {\href{https://proceedings.neurips.cc/paper_files/paper/2020/file/1457c0d6bfcb4967418bfb8ac142f64a-Paper.pdf}{NeurIPS}},
  series    = {Adv. Neural Inf. Process. Syst.},
  volume    = {33},
  pages     = {1877--1901},
  year      = {2020},
  editor    = {Larochelle, Hugo and Ranzato, Marc and Hadsell, Raia and Balcan, Maria-Florina and Lin, Hsuan-Tien},
  publisher = {Curran Associates, Inc.},
  doi       = {10.48550/arXiv.2005.14165},
  note      = {\href{https://arxiv.org/abs/2005.14165}{arXiv:2005.14165}},
}

@inproceedings{dong2024_inContextLearningSurvey,
  author    = {Dong, Qingxiu and Li, Lei and Dai, Damai and Zheng, Ce and Ma, Jingyuan and Li, Rui and Xia, Heming and Xu, Jingjing and Wu, Zhiyong and Chang, Baobao and Sun, Xu and Li, Lei and Sui, Zhifang},
  title     = {A Survey on In-context Learning},
  booktitle = {\href{https://aclanthology.org/2024.emnlp-main.0/}{EMNLP}},
  year      = {2024},
  month     = nov,
  address   = {Miami, Florida, USA},
  publisher = {Association for Computational Linguistics},
  pages     = {1107--1128},
  doi       = {10.18653/v1/2024.emnlp-main.64},
  note      = {\href{https://arxiv.org/abs/2301.00234}{arXiv:2301.00234}},
}

@article{giannakis2021_streamingKAF,
  author        = {Giannakis, Dimitris and Henriksen, Amelia and Tropp, Joel A. and Ward, Rachel},
  title         = {Learning to Forecast Dynamical Systems from Streaming Data},
  year          = {2021},
  month         = sep,
  doi           = {10.48550/arXiv.2109.09703},
  journal       = {\href{https://arxiv.org/abs/2109.09703}{arXiv:2109.09703}},
}

\newpage
\appendix
\onecolumn

\tableofcontents
\newpage

\section{Theory}

\subsection{Nonlinear Laplacian Spectral Analysis \label{NLSA}}

Nonlinear Laplacian Spectral Analysis (NLSA) \cite{Giannakis2012_NLSA} applies the diffusion--map construction of \S\ref{sec:softmax} to delay--embedded time series. Let $R_{tX} \in \mathbb{R}^{N\times D}$ denote a multivariate time-series, and let $\mathscr{H}_L$ be a Hankelization operator of window length $L$ that maps $R$ to a cloud of delay vectors:
\begin{align}\label{takens_embedding}
R_{TcX} = (\mathscr{H}(R_{tX}))_{TcX} = R_{(T+c-1),X} \in \mathbb{R}^{K \times L \times D}\quad.
\end{align}
where $T = \mathbb{Z}_{K}$ indexes the $K = N-L+1$ overlapping windows and $X c$ collects the delayed channel indices. NLSA then builds a diffusion kernel on these delay vectors using the same Gaussian construction as in \S\ref{sec:softmax}. Writing
\begin{align*}
  D^2_{TT'} &= \| R_{T X c} - R_{T' X c} \|_2^2, \\
  P^{+}_{TT'} &= \softmax^{+}_{\alpha}\!\left( - \beta D^2_{TT'} \right),
\end{align*}
we obtain a row--stochastic Markov operator $P^+$ on delay time indices $T$, whose nontrivial eigenvectors $\{R_{Tx}\}_{x\ge 1}$ define nonlinear spatiotemporal modes adapted to the underlying dynamics and sampling density. In LISA we treat this NLSA encoder $R_{tX} \mapsto R_{Tx}$ as a frozen spectral backbone, and introduce an additional in--context ``isocoder'' acting in diffusion space to adapt predictions to each novel time series.

\subsection{Gaussian--Process Latent Model (GPLM) \label{GPLM}}

NLSA provides a nonlinear \emph{encoder} from ambient observations to diffusion coordinates, but forecasting and generation additionally require a \emph{decoder} that maps latent states back to the ambient space. We use a kernel-regression decoder that we refer to as a \emph{Gaussian--Process Latent Model} (GPLM). The terminology is motivated by Gaussian-process regression (GPR), but unlike GP-LVM models the latent variables are treated as fixed inputs (here, diffusion coordinates) rather than optimized parameters~\cite{Rasmussen2005_GP, lawrence2004_gplvm}.

Let $\{(R_{ix}, R_{iX})\}_{i=1}^N$ denote training pairs, where $R_{ix}\in\mathbb{R}^{N\times d}$ are latent diffusion coordinates (e.g., NLSA coordinates of delay windows) and $R_{iX}\in\mathbb{R}^{N\times D}$ are the associated ambient outputs (e.g., next-step targets). We assume a supervised mapping
\begin{align}
    R_{iX} = f(R_{ix}) + \eta_i, \qquad \eta_i \sim \mathcal{N}(0,\sigma^2 I_D),
\end{align}
and place a Gaussian-process prior $f\sim\mathcal{GP}(0,k)$ with an RBF kernel
\begin{align}
    k(R_{ix},R_{jx})
    = \exp\!\left(-\beta\frac{\|R_{ix}-R_{jx}\|_2^2}{\varepsilon}\right),
\end{align}
where $\varepsilon, \beta>0$ is the latent length scale (controlling locality). For a novel latent $R_{ax}$, standard GPR yields a predictive mean and covariance
\begin{align}
    \mathbb{E}[R_{aX}\mid R_{ax}]
    &= k_{a i}\,(K+\sigma^2 I)^{-1} R_{iX}, \label{gplm_full_mean}\\
    \mathrm{Cov}(R_{aX}\mid R_{ax})
    &= \bigl(k_{aa} - k_{a i}(K+\sigma^2 I)^{-1}k_{i a}\bigr)\, I_D, \label{gplm_full_cov}
\end{align}
where $K_{ij}=k(R_{ix},R_{jx})$, $k_{ai}=k(R_{ax},R_{ix})$, and repeated training indices are summed. Equation~\eqref{gplm_full_mean} coincides with the kernel ridge regression (KRR) predictor, while \eqref{gplm_full_cov} provides a principled uncertainty estimate that can be used to monitor confidence during autoregressive rollouts or to generate stochastic samples.

\subsubsection{Autoregression}

To use GPLM for forecasting, we construct training pairs from a time-ordered sequence. Let $x_t\in\mathbb{R}^{D}$ denote the ambient signal and let $z_t\in\mathbb{R}^{d}$ denote the corresponding latent coordinate produced by the NLSA encoder (e.g., the diffusion coordinate of the delay window ending at time $t$). We train the decoder on pairs
\begin{align}
    \{(z_t, x_{t+1})\}_{t=1}^{N-1},
\end{align}
so that the predictive mean \eqref{gplm_full_mean} yields a one-step forecast $\hat{x}_{t+1} = \mathbb{E}[x_{t+1}\mid z_t]$. Multi-step forecasting proceeds autoregressively by iterating the encoder--decoder pipeline: from the current predicted history we form the next latent $z_{t+1}$ and apply GPLM again.

\section{Coifman and Lafon's Softmax \label{sec:softmax}}

Coifman and Lafon's diffusion-map construction \cite{coifman2006_diffusionMaps} can be
understood as a density-corrected softmax that maps a generic logits matrix
$H_{ij}$ to a Markov matrix (In diffusion-maps, one typically takes $H_{ij} = -\frac{D^2_{ij}}{\varepsilon}$, i.e. proportional to the square-distance-matrix).  Starting from $H_{ij}$, define the unnormalized affinity
\begin{equation}
  K_{ij} = \exp(-\beta H_{ij}),
\end{equation}
and the degree (density)
\begin{equation}
  q_i = \sum_j K_{ij},
  \qquad i \in \mathbb{Z}_N,
\end{equation}
with degree matrix $D_q = \mathrm{diag}(q_i)$.  For a parameter
$\alpha \in \mathbb{R}$, the $\alpha$--power symmetric normalization is
\begin{equation}
  K^{(\alpha)}_{ij} =
  \frac{K_{ij}}{q_i^\alpha q_j^\alpha}
  \;=\;
  \bigl(D_q^{-\alpha} K D_q^{-\alpha}\bigr)_{ij}.
  \label{eq:softmax-alpha-symm}
\end{equation}
This step suppresses sampling-density effects for $\alpha>0$ (and reduces to
$K$ when $\alpha=0$).  A row-wise Boltzmann--Gibbs normalization then yields
the Markov operator:
\begin{equation}
  P^+_{ij}
  = \Softmax^+_{\alpha}(-\beta H)_{ij}
  := \frac{K^{(\alpha)}_{ij}}{\sum_k K^{(\alpha)}_{ik}}
  = \frac{\displaystyle \frac{\exp(-\beta H_{ij})}{q_i^\alpha q_j^\alpha}}
         {\displaystyle \sum_k \frac{\exp(-\beta H_{ik})}{q_i^\alpha q_k^\alpha}}.
  \label{eq:softmax-alpha-markov}
\end{equation}
We refer to $H \mapsto P^+$ as the \emph{Coifman--Lafon $\Softmax_\alpha$}. Special cases recover standard operators: $\alpha=0$ gives the random walk on $K$, while $\alpha=1$ yields a diffusion operator whose infinitesimal generator converges (under suitable conditions) to the Laplace--Beltrami operator \cite{coifman2006_diffusionMaps}.  Intermediate values (e.g.\ $\alpha=1/2$) correspond to Fokker--Planck-type normalizations.

\newpage
\begin{sidewaystable*}[hbt!]
\centering
\small
\setlength{\tabcolsep}{5pt}
\renewcommand{\arraystretch}{1.25}
\begin{tabular}{l c c p{5.4cm} c c c p{5.0cm}}
\hline
\textbf{System} & \textbf{Ref.} & \textbf{Dim} & \textbf{Equation} & \textbf{parameters} & $\boldsymbol{dt}$ & $\boldsymbol{L}$ 
\\
\hline

Lorenz--63 & \cite{Lorenz1963} &
3 &
$\begin{aligned}
\dot{x} &= \sigma\,(y - x), \\
\dot{y} &= x(\rho - z) - y, \\
\dot{z} &= xy - \beta z.
\end{aligned}$ & 
$\begin{aligned}
\sigma &=10, \\
\rho &=28,\\
\beta &=8/3
\end{aligned}$ &
0.01 &
75 
\\

Lorenz--96 & \cite{lorenz_predictability_ecmwf_1995} &
5 &
$\begin{aligned}
\dot{x}_i &= (x_{i+1} - x_{i-2})\,x_{i-1} - x_i + F,\\
i&=\mathbb{Z}_K,
\end{aligned}$ &
$\begin{aligned}
K &= 5,\\
F &= 8
\end{aligned}$ &
0.01 &
40 
\\

Rössler &
\cite{rossler1976_continuousChaosEq} &
3 &
$\begin{aligned}
\dot{x} &= -y - z, \\
\dot{y} &= x + a\,y, \\
\dot{z} &= b + z\,(x - c),
\end{aligned}$ & 
$\begin{aligned}
a &= 0.2,\\
b &= 0.2,\\
c &= 5.7
\end{aligned}$ &
0.01 &
1200 
\\

Duffing (NESS) &
\cite{holmes1979_strangeAttractorOscillator} &
4 &
$\ddot x+\delta\dot x+\alpha x+\beta x^3=\gamma\cos(\omega t)$ &
$\begin{aligned}
\alpha &= 0.2,\\
\beta &= 0.2,\\
\gamma &= 5.7, \\
\delta &= 0, \\
\omega &= 2
\end{aligned}$ &
$0.01$ &
628 
\\

Duffing (NE) &
\cite{holmes1979_strangeAttractorOscillator} &
2 &
$\ddot x+\delta\dot x+\alpha x+\beta x^3=\gamma\cos(\omega t)$ &
$\begin{aligned}
\alpha &= 0.2,\\
\beta &= 0.2,\\
\gamma &= 5.7, \\
\delta &= 0, \\
\omega &= 2
\end{aligned}$ &
$0.01$ &
628 
\\

Chua circuit &
\cite{matsumoto1985_doubleScroll} &
3 &
$\begin{aligned}
\dot x &= \alpha(y-x-h(x)+m_1 x),\\
\dot y &= x-y+z,\\
\dot z &= -\beta y \\
h(x)   &= \frac{1}{2} (m_0-m_1)(|x+1|-|x-1|) \\
\end{aligned}$ &
$\begin{aligned}
\alpha &= 15.6,\\
\beta &= 28.0,\\
m_0 &= -1.15,\\
m_1 &= -0.70,
\end{aligned}$ &
$0.005$ &
175 
\\

Halvorsen &
\cite{sprott2003_chaosTimeSeries} &
3 &
$\begin{aligned}
\dot{x} &= -a\,x - 4y - 4z - y^{2}, \\
\dot{y} &= -a\,y - 4z - 4x - z^{2}, \\
\dot{z} &= -a\,z - 4x - 4y - x^{2}.
\end{aligned}$ &
$a=1.4$ &
0.005 &
 
\\

Torus (quasiperiodic) &
\cite{strogatz2018_nonlinearDynamicsChaos} &
3 &
$\begin{aligned}
x&=(R+r\cos\theta_2)\cos\theta_1, \\
y&=(R+r\cos\theta_2)\sin\theta_1,\\
z&=r\sin\theta_2
\end{aligned}$ &
$\begin{aligned}
R&=2, \\
r&=0.7,\\
\omega_1 &=1, \\
\omega_2 &=\sqrt{2}, \\
\end{aligned}$ & 
$0.01$ &

\\

\hline
\end{tabular}
\caption{Reference summary for benchmark dynamical systems.}
\label{tab:attractors_lyapunov_reference}
\end{sidewaystable*}



\newpage
\begin{sidewaystable}[!hbt]
\centering
\small
\setlength{\tabcolsep}{5pt}
\renewcommand{\arraystretch}{1.25}
\begin{tabular}{l l l} 
\toprule
\textbf{Abbrev.} & \textbf{Full name} & \textbf{Equation} 
\\
\midrule

MSE &
Mean Squared Error &
$\displaystyle
\mathrm{MSE}(F,T)=
\frac{1}{A}\sum_{a=1}^{A}\,\big\|F_{a\cdot}-T_{a\cdot}\big\|_2^2
$
\\

Spec-JS (or Spec-KL) &
Spectral Jensen--Shannon (or KL) Divergence &
$\displaystyle
D_{\mathrm{JS}}(p_T\,\|\,p_F)
=
\frac{1}{2}D_{\mathrm{KL}}(p_T\,\|\,m)
+
\frac{1}{2}D_{\mathrm{KL}}(p_F\,\|\,m),
\quad
m=\tfrac{1}{2}(p_T+p_F)
$
\\

ACF-MSE &
Autocorrelation Function MSE &
$\displaystyle
\mathrm{ACF\text{-}MSE}(F,T)=
\frac{1}{\tau_{\max}}\sum_{\tau=1}^{\tau_{\max}}
\frac{1}{X}\sum_{x=1}^{X}
\Big(\widehat{\rho}_{F,x}(\tau)-\widehat{\rho}_{T,x}(\tau)\Big)^2
$
\\

MMD$^2$ &
Squared Maximum Mean Discrepancy &
$\displaystyle
\mathrm{MMD}^2(X_T,X_F)=
\frac{1}{n(n-1)}\!\!\sum_{i\neq i'} k(z_i,z_{i'})
+
\frac{1}{m(m-1)}\!\!\sum_{j\neq j'} k(w_j,w_{j'})
-
\frac{2}{nm}\sum_{i,j} k(z_i,w_j)
$
\\

\bottomrule
\end{tabular}

\caption{
Error metrics used to evaluate dynamical fidelity beyond pointwise prediction accuracy.
The truth and forecast time series are $T_{aX}\in\mathbb{R}^{A\times X}$ and $F_{aX}\in\mathbb{R}^{A\times X}$, where
$a\in\{1,\dots,A\}$ indexes time (forecast steps) and $x\in\{1,\dots,X\}$ indexes coordinates/features; $F_{a\cdot}$ denotes the $X$-vector at time $a$.
For spectral divergence, let $\widehat S_{T,\omega x}$ and $\widehat S_{F,\omega x}$ denote estimated power spectra (e.g.\ Welch or FFT-based periodograms) over discrete frequencies $\omega\in\Omega$ for each coordinate $x$.
We form normalized spectral probability masses
$p_{T}(\omega\mid x)=\frac{\widehat S_{T,\omega x}+\epsilon}{\sum_{\omega'\in\Omega}(\widehat S_{T,\omega' x}+\epsilon)}$
and similarly $p_{F}(\omega\mid x)$, with $\epsilon>0$ for numerical stability (optionally excluding $\omega=0$).
The table reports either Jensen--Shannon divergence $D_{\mathrm{JS}}$ (symmetric, bounded) or KL divergence $D_{\mathrm{KL}}$; in practice we average over coordinates:
$\frac{1}{X}\sum_{x=1}^{X} D_{\mathrm{JS}}\!\big(p_T(\cdot\mid x)\,\|\,p_F(\cdot\mid x)\big)$.
For ACF-MSE, $\widehat{\rho}_{T,x}(\tau)$ and $\widehat{\rho}_{F,x}(\tau)$ are sample autocorrelations of coordinate $x$ at lag $\tau$ up to $\tau_{\max}$.
For MMD$^2$, $X_T=\{z_i\}_{i=1}^n$ and $X_F=\{w_j\}_{j=1}^m$ are samples from the true and predicted processes (e.g.\ states or delay vectors), and $k(\cdot,\cdot)$ is a positive-definite kernel (e.g.\ RBF).
}
\label{tab:error-metrics}
\end{sidewaystable}


\end{document}